\documentclass[10pt,twocolumn,letterpaper]{article}

\usepackage{cvpr}

\definecolor{cvprblue}{rgb}{0.21,0.49,0.74}
\usepackage[pagebackref,breaklinks,colorlinks,allcolors=cvprblue]{hyperref}
\usepackage{bm}

\def\paperID{6}
\def\confName{CVPR}
\def\confYear{2026}

\title{SPG: Sparse-Projected Guides with Sparse Autoencoders for Zero-Shot Anomaly Detection}

\author{Tomoyasu Nanaumi$^{1}$\thanks{Corresponding author: tm-nanaumi@yachiyo-eng.co.jp}\quad Yukino Tsuzuki$^{1}$\quad Junichi Okubo$^{1}$\quad Junichiro Fujii$^{1}$\quad Takayoshi Yamashita$^{2}$\\
$^{1}$ Yachiyo Engineering Co., Ltd., Japan \\
$^{2}$ Chubu University, Japan \\
}

\begin{document}
\maketitle

\begin{abstract}
We study zero-shot anomaly detection and segmentation using frozen foundation model features, where all learnable parameters are trained only on a labeled auxiliary dataset and deployed to unseen target categories without any target-domain adaptation.
Existing prompt-based approaches use handcrafted or learned prompt embeddings as reference vectors for normal/anomalous states.
We propose Sparse-Projected Guides~(SPG), a prompt-free framework that learns sparse guide coefficients in the Sparse Autoencoder~(SAE) latent space, which generate normal/anomaly guide vectors via the SAE dictionary. 
SPG employs a two stage learning strategy on the labeled auxiliary dataset: (i) train an SAE on patch-token features, and (ii) optimize only guide coefficients using auxiliary pixel-level masks while freezing the backbone and SAE.
On MVTec AD and VisA under cross-dataset zero-shot settings, SPG achieves competitive image-level detection and strong pixel-level segmentation; with DINOv3, SPG attains the highest pixel-level AUROC among the compared methods.
We also report SPG instantiated with OpenCLIP~(ViT-L/14@336px) to align the backbone with CLIP-based baselines.
Moreover, the learned guide coefficients trace decisions back to a small set of dictionary atoms, revealing category-general and category-specific factors.
\end{abstract}

\section{Introduction}
\label{sec:intro}

Industrial visual inspection requires anomaly detection and anomaly segmentation,  i.e., deciding whether an image is defective and localizing the defective regions.
A wide range of approaches has been studied for this task, including formulations as binary (normal vs. anomalous) classification~\cite{baitieva_2024_CVPR_SegAD} as well as one-class learning that trains only on normal data~\cite{Roth_2022_CVPR_PatchCore}.
Among these, student–teacher-based approaches have achieved strong performance on popular benchmarks. For example, UniNet~\cite{Wei_2025_CVPR_UniNet} reports 99.0\% AUROC on MVTec AD~\cite{Bergmann_2019_CVPR_MVTec} and 98.9\% AUROC on VisA~\cite{Zou_2022_ECCV_VisA}. However, many of these methods implicitly assume that training data can be collected for each target category. 
This limitation has increased interest in zero-shot anomaly detection, where target categories are unseen during training, with no target-domain adaptation.

To make zero-shot anomaly detection feasible, a common strategy is to rely on features from foundation models pretrained on large-scale datasets~(e.g., CLIP~\cite{Radfold_2021_ICML_CLIP}, DINO~\cite{simeoni_2025_dinov3}). 
Such representations often encode generic visual cues that can separate normal from anomalous patterns even when the target categories are unseen during training.
Building on this idea, CLIP-based zero-shot methods separate normal and anomalous states using handcrafted or learned prompts~\cite{Jeong_2023_CVPR_WinCLIP, Zhou_2024_ICLR_AnomalyCLIP}. 
While effective, these approaches typically rely on a vision–language backbone via text prompting. In contrast, we seek a prompt-free criterion defined directly on frozen visual features.

To this end, we propose Sparse-Projected Guides~(SPG), a prompt-free framework for zero-shot anomaly detection that uses guide vectors as an alternative to prompts. Our design is inspired by recent progress on Sparse Autoencoders (SAEs)~\cite{makhzani_2013_ksparse}, 
which map representations inside foundation models into a sparse, structured latent space over a learned dictionary~\cite{bricken_2023_monosemanticity, gao_2025_ICLR_scalingSAE}. 
Building on this insight, we construct guide vectors via an SAE-induced sparse latent space and propose a tunable and analyzable framework for zero-shot anomaly detection and segmentation. 

We summarize our main contributions as follows:

\begin{itemize}
    \item \textbf{Prompt-free, backbone-agnostic framework.} We introduce SPG, a prompt-free framework for zero-shot anomaly detection and segmentation that defines normal/anomaly guide vectors via an SAE dictionary and its sparse latent space. Since SPG uses only frozen patch-token features, it is compatible with both VLM and vision-only backbones (e.g., CLIP and DINOv3).
    
    \item \textbf{Competitive detection and strong segmentation.} In cross-dataset zero-shot settings, SPG achieves competitive image-level detection and strong pixel-level segmentation. In particular, SPG with DINOv3 achieves the highest pixel-level AUROC in our comparisons on both MVTec AD and VisA.
    \item \textbf{Ablations on key design choices.} We analyze how SPG’s performance depends on three factors: (i) SAE dictionary capacity/sparsity, (ii) the frozen feature backbone, and (iii) the aggregation rule for image-level scoring.
    \item \textbf{Findings on class-general vs. class-specific factors.}  By analyzing the SAE latent space, we show that class-general factors (e.g., cracks and missing parts) coexist with class-specific factors.
\end{itemize}

\section{Related Works}
\label{sec:relatedworks}

\subsection{Foundation-Model-based Anomaly Detection}
Foundation models have recently become a common backbone for anomaly detection and segmentation, enabling decision criteria to be defined in frozen representation spaces. 

Zero-shot anomaly detection and segmentation (ZSAD) is one such setting, where target categories are unseen during training and no target-domain adaptation is performed.
A common approach uses vision–language models (VLMs) such as CLIP~\cite{Radfold_2021_ICML_CLIP}: text embeddings representing normal and abnormal concepts (i.e., prompts) serve as reference vectors, and similarity between image features and these references yields both image-level decisions and pixel-level segmentation maps.
WinCLIP~\cite{Jeong_2023_CVPR_WinCLIP}, a pioneering CLIP-based ZSAD method, combines compositional prompt ensembles with window-level feature aggregation for anomaly classification and segmentation.
Subsequent studies largely move toward auxiliary-data-driven prompt learning or representation adaptation, including object-agnostic prompts (AnomalyCLIP~\cite{Zhou_2024_ICLR_AnomalyCLIP}), hybrid static/dynamic prompts (AdaCLIP~\cite{Cao_2024_ECCV_AdaClip}), visual-context prompting (VCP-CLIP~\cite{qu_2024_ECCV_vcpclip}), anomaly-aware text anchors with patch alignment (AA-CLIP~\cite{ma_2025_CVPR_aaclip}), and adapter-based adaptation for scoring and localization (AdaptCLIP~\cite{gao_2026_AAAI_adaptclip}).
Overall, CLIP-based ZSAD has largely been driven by prompt-centric designs, either training-free prompt engineering or auxiliary-data-driven prompt/representation adaptation.

Beyond CLIP-style prompting, AnomalyDINO~\cite{damm_2025_WACV_anomalydino} does not rely on textual prompts and performs anomaly detection and segmentation using frozen features from a self-supervised visual foundation model (DINOv2~\cite{oquab_2024_TMLR_dinov2}).
While it is primarily evaluated in one-/few-shot settings, it highlights that strong frozen visual backbones can serve as effective representations for anomaly localization, motivating prompt-free criteria that operate purely on visual patch tokens.

Prior work largely defines normal/anomalous reference vectors using CLIP prompts (training-free or auxiliary-adapted), whereas vision-only backbones such as DINO have enabled exploration of prompt-free alternatives under different evaluation settings. 
We propose SPG, a prompt-free framework that constructs normal/anomaly guide vectors as references via an SAE dictionary and its sparse latent space.

\subsection{Sparse Autoencoders}
Sparse autoencoders (SAEs) are a promising approach in mechanistic interpretability for decomposing superposed representations into sparse linear combinations over a learned dictionary~\cite{makhzani_2013_ksparse, gao_2025_ICLR_scalingSAE, bricken_2023_monosemanticity}. 
However, training SAEs is often non-trivial in practice.
Recovering a large number of concepts often requires an overcomplete and wide dictionary, which makes balancing reconstruction error and sparsity difficult. Moreover, training can produce many dead latents (units that rarely activate), reducing the effective dictionary capacity. 
These issues are frequently highlighted as becoming more pronounced when scaling SAEs.

To address such challenges, recent work has proposed training schemes that directly control sparsity, such as TopK (k-sparse) SAEs, and has provided frameworks to organize scaling behavior and evaluation metrics while mitigating dead latents even with wide dictionaries~\cite{gao_2025_ICLR_scalingSAE}.

Building on these insights, we train and analyze an SAE over visual patch-token features of a frozen image encoder, and define a ZSAD framework that constructs normal/anomaly guide vectors as reference vectors via the SAE dictionary and its sparse latent space. 

\section{Method}
\label{sec:method}

\begin{figure*}
    \centering
    \includegraphics[width=1.0\linewidth, trim=0 20 20 20]{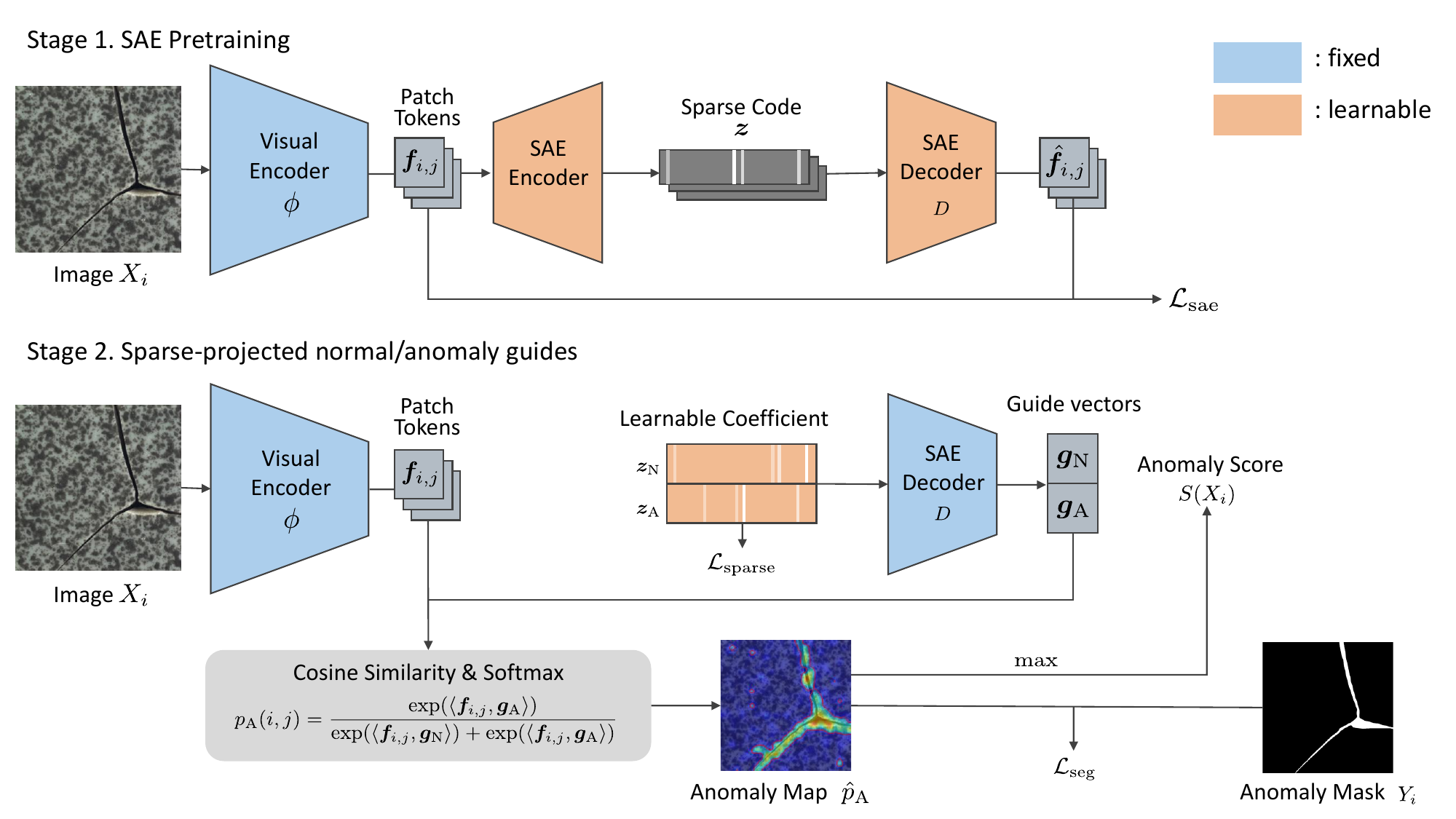}
    \caption{Overview of SPG~(Sparse-Projected Guides). Stage 1 trains a Sparse Autoencoder~(SAE) on patch-token features extracted by a frozen visual encoder to learn a dictionary in which normal/anomaly guide vectors can be parameterized as sparse coefficients. Stage 2 freezes both the encoder and SAE, and learns only non-negative, sparsity-regularized guide coefficients that generate normal/anomaly guide vectors via the SAE dictionary. At inference, cosine similarities between patch tokens and the two guides are converted into an anomaly probability map via a two-class softmax, which is upsampled to the image resolution. Image-level anomaly scores are computed by aggregating the anomaly map~(default: max pooling; see \cref{equ:lse_pooling} for the log-sum-exp generalization with temperature $\tau$).}
    \label{fig:overview}
\end{figure*}

\subsection{Overview}
We propose a two-stage learning framework on the frozen feature space of a pretrained visual encoder, as shown in \cref{fig:overview}. We parameterize normal/anomaly guide vectors by sparse coefficients in the SAE latent space.
Stage 1 learns an SAE on patch-token features to obtain a dictionary that enables sparse decomposition of token representations.
Stage 2 then freezes the SAE and learns the guide coefficients in the SAE latent space. The learned guide coefficients are transferred to a target dataset without any additional training, and the corresponding guide vectors are computed via the SAE dictionary.

\subsubsection{Problem setting.}
We study zero-shot anomaly detection and segmentation: we learn from labeled training data and deploy on unseen target categories without any target-domain adaptation.

We denote a labeled auxiliary (training) dataset and a target dataset as
$
\mathcal{D}_{\mathrm{train}}=\{(X_i, Y_i, y_i)\}_{i=1}^{N_{\mathrm{train}}},\quad
\mathcal{D}_{\mathrm{target}}=\{(X_i, Y_i, y_i)\}_{i=1}^{N_{\mathrm{target}}}.
$
Here, $X_i\in\mathbb{R}^{H\times W\times 3}$ is an image, $Y_i\in\{0,1\}^{H\times W}$ is a pixel-level anomaly mask, $y_i\in\{0,1\}$ is an image-level anomaly label. 
The categories of $\mathcal{D}_{\mathrm{train}}$ and $\mathcal{D}_{\mathrm{target}}$ are disjoint. 
All learnable parameters are trained only on $\mathcal{D}_{\mathrm{train}}$, and the resulting model is applied to $\mathcal{D}_{\mathrm{target}}$ without further training. 
In practical inference on the target domain, $Y_i$ and $y_i$ are not used and are referenced only for evaluation.

\subsubsection{Frozen visual encoder and token features.}
We use a pretrained visual encoder $\phi$ (e.g., the visual backbone of VLMs or self-supervised ViTs) and keep it frozen throughout all stages.
Given an image $X_i$, the encoder outputs a sequence of patch-token features 
\begin{equation}\label{equ:feature_extraction}
    \phi(X_i) = \{\bm{f}_{i,j}\}_{j=1}^{m}, \quad \bm{f}_{i,j} \in \mathbb{R}^d ,
\end{equation}
where $m = H_p W_p$ is the number of patches on an $H_p \times W_p$ patch grid.

\subsubsection{Sparse autoencoder (SAE) as a sparse dictionary representation.}
For each token feature $\bm f\in\mathbb{R}^d$, the SAE maps it to a higher-dimensional intermediate representation $\bar{\bm z}\in\mathbb{R}^C (C \gg d)$, applies a sparsification operator $\mathcal S(\cdot)$ to obtain a sparse code $\tilde{\bm z}\in\mathbb{R}^C$, and reconstructs the original feature using a linear dictionary $D=[\bm d_1,\dots,\bm d_C]\in\mathbb{R}^{d\times C}$:
\begin{equation}\label{equ:sae}
    \bar{\bm z}=\mathrm{Enc}(\bm f),\quad
    \tilde{\bm z}=\mathcal S(\bar{\bm z}),\quad
    \hat{\bm f}=\mathrm{Dec}(\tilde{\bm z})=D\tilde{\bm z},
\end{equation}
where $\mathrm{Enc}(\cdot)$ is a single linear layer that maps from $\mathbb{R}^d$ to $\mathbb{R}^C$.
Common choices for $\mathcal S(\cdot)$ include (i) enforcing non-negativity (e.g., ReLU) with $\ell_1$ regularization to obtain additive sparse codes, or (ii) explicit $k$-sparse projection such as TopK, which keeps the $k$ largest activations and sets the rest to zero. 
In this work, we adopt TopK.

\subsection{Stage 1: SAE pretraining on patch tokens.}
In Stage 1, we keep the visual encoder $\phi$ fixed and train the SAE on patch tokens from $\mathcal D_{\mathrm{train}}$. 
The goal is to represent token features as sparse linear combinations of an overcomplete dictionary.

For each patch token feature $\bm f_{i,j}$ extracted from image $X_i$, we compute an intermediate code and apply TopK sparsification:
\begin{equation}\label{equ:sae_encode}
    \bar{z_{i,j}} = \mathrm{Enc}(\bm{f}_{i,j}), \quad \tilde{\bm{z}}_{i,j}=\mathrm{TopK}(\bar{z}_{i,j}; k)
\end{equation}
where 
$\mathrm{TopK}(\cdot ;k)$ keeps only the top-$k$ entries and sets the others to zero. Reconstruction is given by $\hat{\bm{f}} = D\tilde{z}_{i,j}$. We train the SAE by minimizing the reconstruction loss:
\begin{equation}
    \mathcal{L}_{\mathrm{sae}} = \frac{1}{N_{\mathrm{train}}m}\sum_{i=1}^{N_{\mathrm{train}}}\sum_{j=1}^{m}\| \bm{f}_{i,j} - D\tilde{z}_{i,j} \|^2_2.
\end{equation}
TopK-based sparsification explicitly controls the sparsity level compared to $\ell_1$-regularized alternatives, which can improve training stability. After Stage 1, the learned dictionary 
$D$ and encoder 
$\mathrm{Enc}$ are frozen and are not updated in later stages.

\subsection{Stage 2: Sparse-projected normal/anomaly guide vectors in dictionary coordinates.}
In Stage 2, we learn guide coefficients in the SAE latent space, which define normal/anomaly guide vectors via the SAE dictionary.
While prior methods often use handcrafted or learned prompt embeddings as reference vectors, our guides are expressed as sparse linear combinations of SAE dictionary atoms, which aims to improve interpretability and controllability.

\noindent\textbf{Parameterization of guide coefficients.}
We introduce learnable guide coefficient parameters for the normal and anomaly guide vectors:
\begin{equation}
    \bar{z}_N, \bar{z}_A \in \mathbb{R}^C,
\end{equation}

and apply an element-wise non-negativity operator $\rho(\cdot)$ (e.g., ReLU or softplus) to obtain non-negative coefficients:
\begin{equation}\label{equ:sparsify_guide_code}
    \tilde{z}_N = \rho(\bar{z}_N), \quad \tilde{z}_A = \rho(\bar{z}_A).
\end{equation}
The corresponding guide vectors in the token feature space are defined by the SAE dictionary:
\begin{equation} \label{equ:guide}
    \bm{g}_N = D\tilde{z}_N, \quad \bm{g}_A = D\tilde{z}_A.
\end{equation}

Let $\mathrm{supp}(\tilde{z})$ denote the indices of non-zero coefficients. We define the active atom sets as
\begin{equation}\label{equ:active_set}
    \mathcal{I}_N = \mathrm{supp}(\tilde{z}_N), \quad \mathcal{I}_A = \mathrm{supp}(\tilde{z}_A).
\end{equation}
Because we impose sparsity regularization on $\tilde{z}_N, \tilde{z}_A$~, these sets remain small. This property is exploited in \cref{subsec:vis_atoms} to visualize which dictionary atoms each guide vector emphasizes.

\noindent\textbf{Anomaly Scoring.}
For each patch token feature $\bm{f}_{i,j}$, we $\ell_2$-normalize both $\bm{f}_{i,j}$ and the guide vectors and compute cosine similarities to obtain two logits:
\begin{equation}
    \ell_N (i,j) = \cos(\bm{f}_{i,j}, \bm{g}_N), \quad \ell_A (i,j) = \cos(\bm{f}_{i,j}, \bm{g}_A),
\end{equation}
We then obtain the anomaly probability via a temperature-scaled two-class softmax (unless otherwise stated, $T=0.07$):
\begin{equation}
    p_A(i,j) = \frac{\exp(\ell_A (i,j)/T)}{\exp(\ell_N (i,j)/T) + \exp(\ell_A (i,j)/T)}.
\end{equation}
We then upsample it to the input image resolution (bilinear interpolation):
\begin{equation} \label{equ:upsample}
    \hat{p}_A = \mathrm{Upsample}(p_A) \in [0,1]^{H \times W}.
\end{equation}
To obtain an image-level anomaly score, we aggregate the anomaly map $\hat{p}_A$ with temperature-controlled log-sum-exp, which smoothly interpolates between max- and mean-pooling:
\begin{equation} \label{equ:lse_pooling}
    S_\tau(X_i) = \tau \log \left( \frac{1}{HW} \sum_{u,v} \exp \left( \frac{\hat{p}_A (u,v)}{\tau} \right) \right).
\end{equation}
Smaller $\tau$ makes the aggregation max-like ($\tau \!\to\! 0$), whereas larger $\tau$ aggregates evidence more uniformly (mean-like); we use max pooling by default and analyze $\tau$ in \cref{subsec:quant_eval}.

\noindent\textbf{Training.}
In Stage 2, we learn the guide coefficients using pixel-level anomaly masks $Y$ from $\mathcal{D}_{\mathrm{train}}$ as supervision. The loss is the sum of a segmentation loss and a sparsity regularizer:
\begin{equation}
    \mathcal{L} = \mathcal{L}_{\mathrm{seg}}(\hat{p}_A, Y) + \mathcal{L}_{\mathrm{sparse}}(\tilde{z}_N, \tilde{z}_A).
\end{equation}
We use a combination of focal loss~\cite{Lin_2017_ICCV_FocalLoss} and Dice loss~\cite{Milletari_2016_3DV_DiceLoss} for segmentation:
\begin{equation} \label{equ:seg_loss}
    \mathcal{L}_{\mathrm{seg}} (\hat{p}_A, Y) = \mathcal{L}_{\mathrm{focal}}(\hat{p}_A, Y)+\lambda_{\mathrm{dice}}\mathcal{L}_{\mathrm{dice}}(\hat{p}_A, Y)
\end{equation}
To encourage sparsity, we apply an $\ell_1$ penalty to the guide coefficients:
\begin{equation} \label{equ:saprsity_loss}
    \mathcal{L}_{\mathrm{sparse}} = \beta (\|\tilde{z}_N\|_1 + \|\tilde{z}_A\|_1)
\end{equation}

Therefore, Stage 2 optimizes only the guide coefficient parameters while keeping the visual encoder and SAE fixed:
\begin{equation}
    \min_{\bar{z}_N, \bar{z}_A} \mathcal{L}_{\mathrm{seg}}(\hat{p}_A, Y) + \beta (\|\rho(\bar{z}_N)\|_1 + \|\rho(\bar{z}_A)\|_1).
\end{equation}
This design enables cross-dataset transfer: the guide vectors learned on the auxiliary dataset can be applied to the target dataset without additional training.

\section{Experiments}

\subsection{Experimental Setup}
\label{subsec:setup}

\noindent\textbf{Datasets and protocol.}
We use MVTec AD~\cite{Bergmann_2019_CVPR_MVTec} and VisA~\cite{Zou_2022_ECCV_VisA}. Following prior work, we evaluate in a cross-dataset zero-shot setting: one dataset is treated as an auxiliary dataset, and the other as a target dataset.

\noindent\textbf{Implementation details.}
We use DINOv3~\cite{simeoni_2025_dinov3}~(ViT-L)\footnote{\url{https://huggingface.co/facebook/dinov3-vitl16-pretrain-lvd1689m}} as the default image encoder. We also report results with OpenCLIP~(ViT-L/14@336px)\footnote{\url{https://github.com/mlfoundations/open_clip}} to compare with prior CLIP-based methods.
All images are resized to 448×448 for DINOv3 and 518×518 for OpenCLIP, and patch-token features are extracted from the final~(24th) transformer block, excluding the CLS token.
For both Stage 1 and Stage 2, we use the Adam~\cite{kingma_2015_ICLR_adam} optimizer with learning rate 0.001 and batch size 16.

\noindent\textbf{Stage 1: SAE pretraining.}
In Stage 1, we train an SAE on token features from the auxiliary dataset to obtain the dictionary vectors.
We train for 50 epochs on the auxiliary dataset.
The width of dictionary is set to $C=4096$.
We use a TopK sparsification operator $\mathcal{S}=\mathrm{TopK}$ with $k=32$. 

\noindent\textbf{Stage 2: Sparse-projected normal/anomaly guide vectors.}
In Stage 2, we learn guide coefficients (normal and anomaly) in the SAE latent space, and construct the corresponding guide vectors via the frozen SAE dictionary from Stage 1.
To stabilize training, we apply exponential moving average (EMA) to the guide coefficient parameters $(\bar{z}_N, \bar{z}_A)$ with a decay coefficient of 0.999.
We use ReLU as the non-negativity operator $\rho (\cdot)$ in \cref{equ:sparsify_guide_code}.
We set $\lambda_{\mathrm{dice}}=1.0$ in \cref{equ:seg_loss} and $\beta=0.01$ in \cref{equ:saprsity_loss} for the sparsity regularizer.

\noindent\textbf{Metrics.}
We quantitatively evaluate both image-level detection and pixel-level segmentation.
For image-level evaluation, we report Area Under the Receiver Operating Characteristic curve~(AUROC) and Average Precision~(AP).
For pixel-level evaluation, we report pixel-level AUROC and AUPRO~\cite{Bergmann_2019_CVPR_MVTec}.

\subsection{Quantitative Evaluation}
\label{subsec:quant_eval}

\begin{table}[t]
\centering
\small
\caption{Cross-dataset zero-shot performance on MVTec AD and VisA (\%). Models are trained on only the other dataset as the unseen target, following the cross-dataset zero-shot protocol with no target-domain adaptation. We report image-level AUROC / AP and pixel-level AUROC / AUPRO~(higher is better). Baseline results are taken from the original papers, and "--" denotes metrics not reported. Note that we report SPG with OpenCLIP to match the backbone used by prior CLIP-based methods, and with DINOv3 ViT-L to demonstrate that SPG is not restricted to VLM backbones.}
\label{tab:sota_comparison}
\setlength{\tabcolsep}{5pt}
\begin{tabular}{lrr|rr}
\hline
\multicolumn{5}{c}{\textbf{Image-level (Detection)}} \\
\hline
Method & \multicolumn{2}{c|}{VisA$\rightarrow$MVTec} & \multicolumn{2}{c}{MVTec$\rightarrow$VisA} \\
& AUROC & AP & AUROC & AP \\
\hline
WinCLIP      & 91.8 & 96.5 & 78.1 & 81.2 \\
AnomalyCLIP  & 91.5 & 96.2 & 82.1 & 85.4 \\
AdaCLIP      & 89.2 & --   & 85.8 & --   \\
VCP-CLIP     & 92.1 & 96.9 & 83.8 & 87.6 \\
AA-CLIP      & 90.5 & --   & 84.6 & --   \\
AdaptCLIP    & 93.5 & 96.7 & 84.8 & 87.6 \\
\hline
SPG (OpenCLIP)   & 79.3 & 89.6 & 82.6 & 85.7 \\
SPG (DINOv3) & 91.4 & 95.7 & 80.2 & 84.3 \\

\hline\hline
\multicolumn{5}{c}{\textbf{Pixel-level (Segmentation)}} \\
\hline
Method & \multicolumn{2}{c|}{VisA$\rightarrow$MVTec} & \multicolumn{2}{c}{MVTec$\rightarrow$VisA} \\
& AUROC & AUPRO & AUROC & AUPRO \\
\hline
WinCLIP      & 85.1 & 64.6 & 79.6 & 56.8 \\
AnomalyCLIP  & 91.1 & 81.4 & 95.5 & 87.0 \\
AdaCLIP      & 88.7 & --   & 95.5 & --   \\
VCP-CLIP     & 92.0 & 87.3 & 95.7 & 90.7 \\
AA-CLIP      & 91.9 & --   & 95.5 & --   \\
AdaptCLIP    & 90.9 & --   & 95.7 & --   \\
\hline
SPG (OpenCLIP) & 89.9 & 71.9 & 94.7 & 87.1 \\
SPG (DINOv3) & 92.3 & 87.7 & 96.0 & 89.1 \\
\hline
\end{tabular}
\end{table}

\begin{figure*}
    \centering
    \includegraphics[width=1.0\linewidth, trim=0 50 0 0 ]{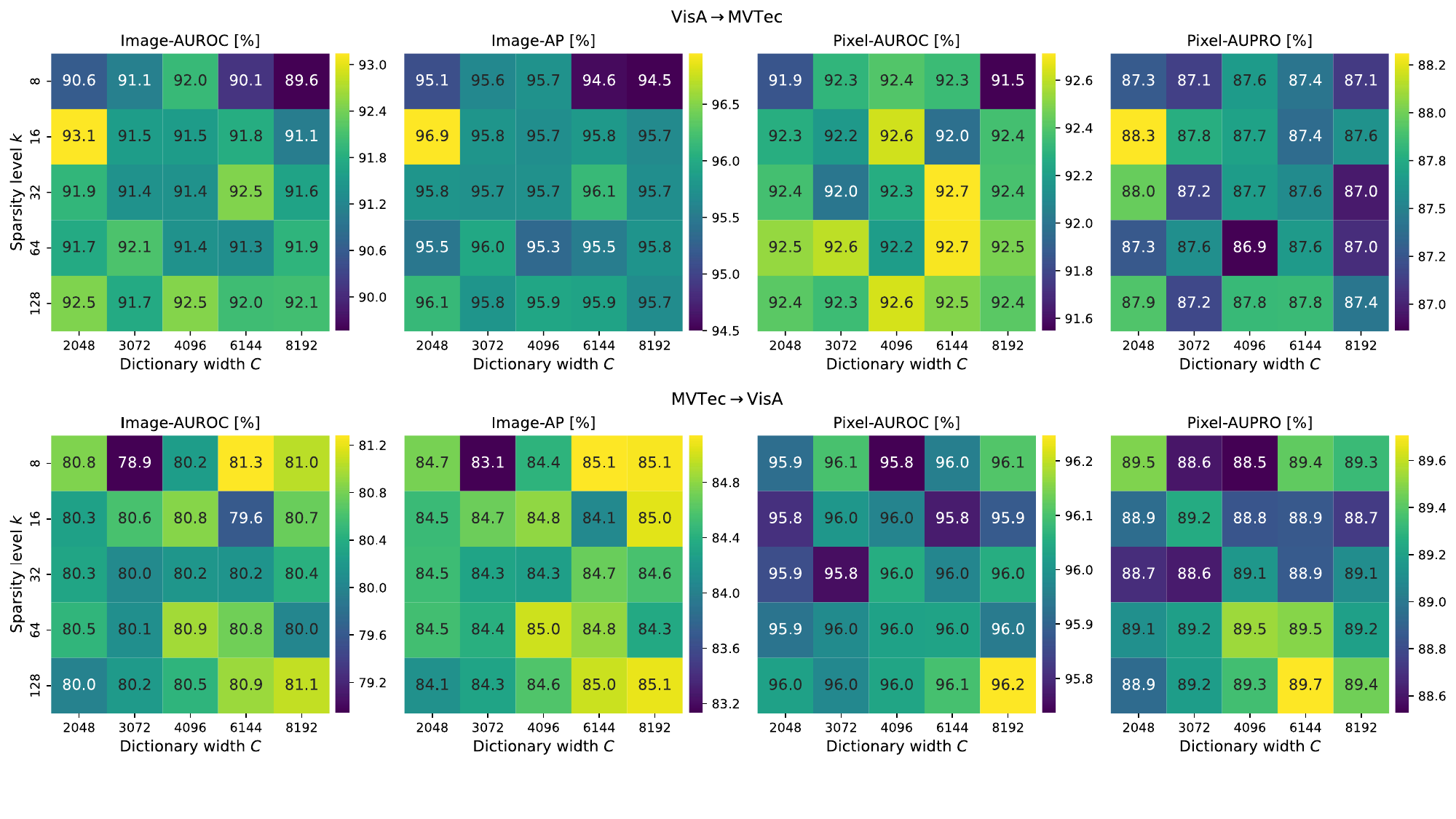}
    \caption{Sensitivity to SAE hyperparameters. The Heatmaps summarize cross-dataset performance when sweeping the SAE dictionary width $C$ and the TopK sparsity level $k$. The top row reports VisA $\rightarrow$ MVTec AD transfer and the bottom row reports MVTec $\rightarrow$ VisA transfer. Each column corresponds to image-level AUROC / AP and pixel-level AUROC / AUPRO, respectively, highlighting that the optimal $(C, k)$ can depend on both the transfer direction and the evaluation metric.}
    \label{fig:ablation_sae_params}
\end{figure*}

\begin{figure}
    \centering
    \includegraphics[width=1.0\linewidth, trim=0 0 400 0]{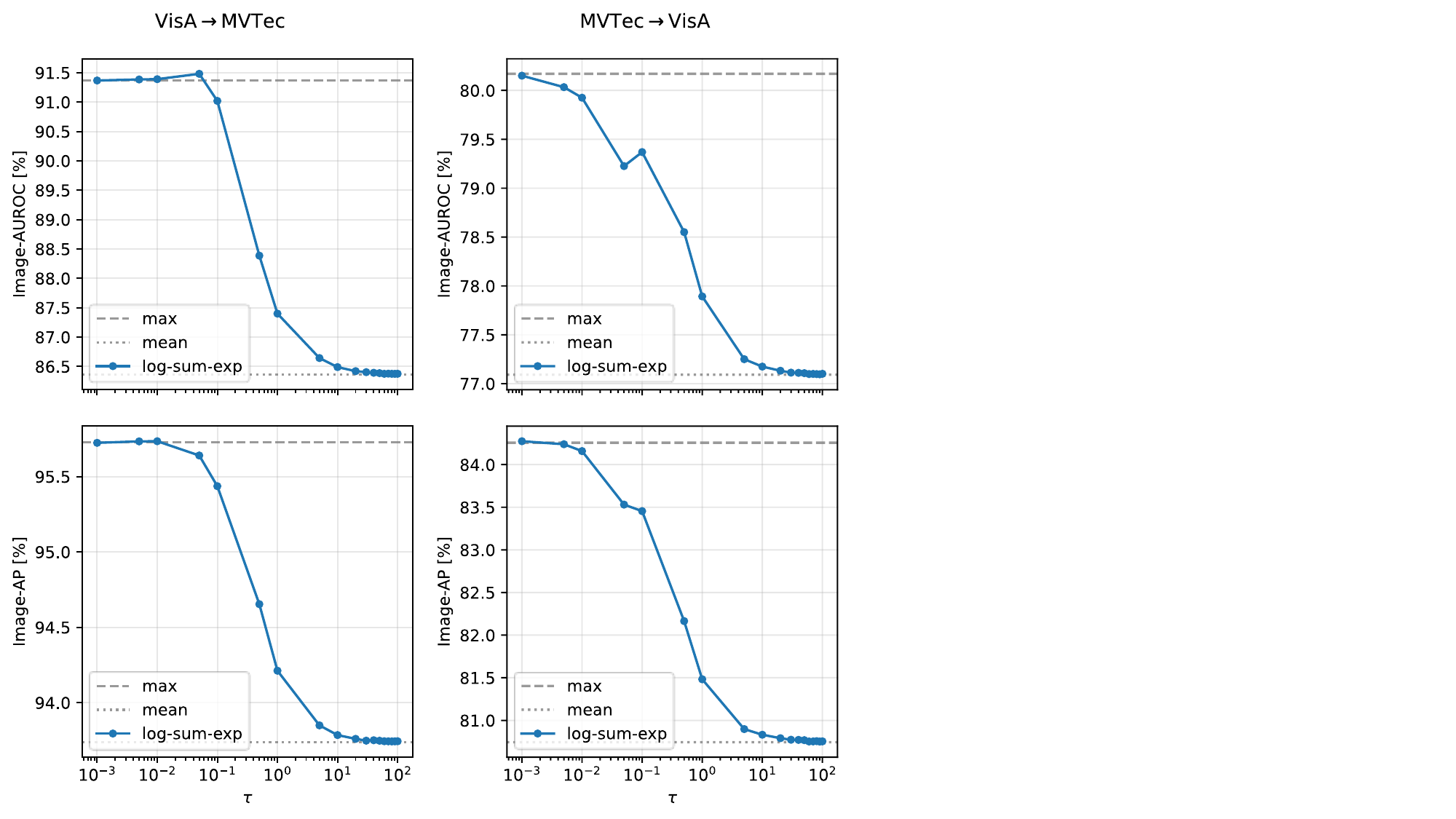}
    \caption{Ablation of image-level anomaly-score aggregation via temperature-controlled log-sum-exp pooling. We vary the temperature $\tau$ in log-sum-exp pooling applied to the upsampled anomaly map to produce an image-level anomaly score. Curves report image-level AUROC and AP for both transfer directions (left: VisA $\rightarrow$ MVTec; right: MVTec $\rightarrow$ VisA). The figure compares the $\tau \rightarrow 0$ (max-like) regime against larger $\tau$ (mean-like) aggregation, showing that more max-like pooling better preserves localized high-confidence anomaly responses for detection. }
    \label{fig:ablation_aggregation}
\end{figure}

We compare SPG with recent zero-shot anomaly detection and segmentation methods on MVTec AD and VisA under the cross-dataset setting. \Cref{tab:sota_comparison} reports image-level AUROC and AP and pixel-level AUROC and AUPRO.

At the image level, our method achieves competitive performance on both datasets. On MVTec AD, it achieves 91.4\% AUROC and 95.7\% AP, comparable to recent CLIP-based approaches. On VisA, the performance (80.2\% AUROC / 84.3\% AP) remains competitive, although some prompt-adaptation methods report higher scores. At the pixel level, our method achieves 92.3\% pixel-level AUROC / 87.7\% AUPRO on MVTec AD and 96.0\% pixel-level AUROC / 89.1\% AUPRO on VisA. Notably, our method yields the best pixel-level AUROC on both MVTec AD and VisA among the compared methods, while its VisA AUPRO is slightly below the strongest prompt-adaptation baseline (e.g., VCP-CLIP).

To better understand which components drive these results and how robust the framework is in practice, we next analyze its sensitivity to several design choices. Specifically, we study:
(i) SAE hyperparameters that control the dictionary capacity and sparsity of the codes,
(ii) the choice of frozen visual backbone $\phi$ used to extract patch tokens,
and (iii) the aggregation function used to convert the patch-level anomaly map into an image-level anomaly score.
Unless otherwise stated, we keep all training schedules and losses fixed and change only the factor under study.

\noindent\textbf{SAE hyperparameter selection.}
We evaluate the effect of the SAE dictionary width $C$ and the TopK sparsity level $k$ on cross-dataset zero-shot performance.
Following \cref{subsec:setup}, we sweep $k \in \{8, 16, 32, 64, 128\}$ and $C \in \{2048, 3072, 4096, 6144, 8192\}$, while keeping all other components fixed. 
\Cref{fig:ablation_sae_params} summarizes the results for both transfer directions and metrics. Overall, the best $(C,k)$ is both metric- and direction-dependent.
For VisA$\rightarrow$MVTec, image-level AUROC/AP peaks at $(C,k)=(2048,16)$, while the pixel-level optimum is less consistent across metrics: pixel-level AUROC slightly increases with a wider dictionary around $C=6144$ and $k=32$--$64$, whereas pixel-level AUPRO attains its maximum at $(2048,16)$.
For MVTec$\rightarrow$VisA, the same $(2048,16)$ setting is not optimal, and pixel-level performance stays close across a broad range of $(C,k)$.
Overall, these results suggest that tuning $(C,k)$ for a single metric or transfer direction may not generalize.

\noindent\textbf{Aggregation for image-level anomaly scoring.}
Our default image-level score uses max pooling over the upsampled anomaly map (\cref{equ:upsample}), which corresponds to the $\tau\rightarrow0$ limit of temperature-controlled log-sum-exp pooling. \cref{fig:ablation_aggregation} compares max/mean pooling with log-sum-exp pooling across temperatures $\tau$. In both auxiliary$\rightarrow$target directions, smaller $\tau$ (i.e., a more “max-like” aggregation) consistently improves image-level AUROC/AP, while larger $\tau$ approaches mean pooling and degrades performance. This suggests that image-level decisions benefit from preserving localized high-confidence responses in the anomaly map; in contrast, uniform averaging can dilute small or spatially sparse anomalies.

\begin{figure}
    \centering
    \includegraphics[width=1.0\linewidth, trim=0 0 580 0]{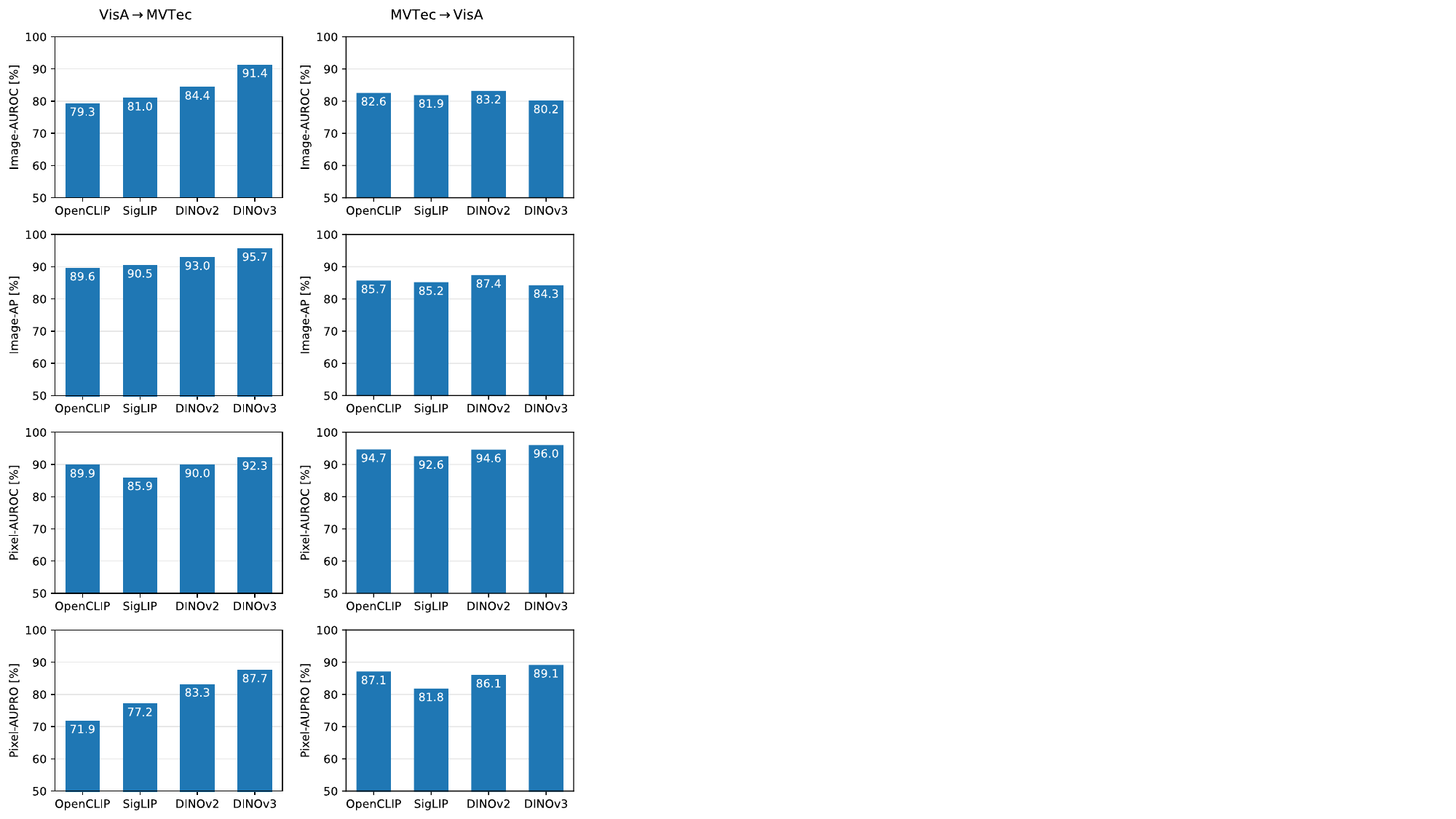}
    \caption{Effect of the visual backbone on SPG under cross-dataset transfer. We instantiate SPG with different frozen encoders and rerun Stage 1 (SAE training on auxiliary patch tokens) and Stage 2 (guide-coefficient learning) for each backbone under the same protocol. Results are reported for two transfer directions (left: VisA $\rightarrow$ MVTec; right: MVTec $\rightarrow$ VisA). Each plot reports image-level AUROC/AP and pixel-level AUROC/AUPRO, isolating how backbone choice influences both detection and segmentation quality within the same guide-based scoring rule. }
    \label{fig:ablation_backbone}
\end{figure}

\noindent\textbf{Effect of backbone choice.}
We analyze how the choice of visual backbone affects SPG under the same training and evaluation protocol.
Specifically, we compare ViT-L backbones from OpenCLIP, SigLIP\footnote{\url{https://huggingface.co/google/siglip-large-patch16-384}}~\cite{Zhai_2023_ICCV_siglip}, DINOv2\footnote{\url{https://huggingface.co/facebook/dinov2-large}}, and DINOv3 by rerunning Stage~1 (SAE training on auxiliary patch tokens) and Stage~2 (guide coefficient learning) for each backbone.
\Cref{fig:ablation_backbone} summarizes the results for two cross-dataset settings.
For VisA$\rightarrow$MVTec, DINOv3 yields the best performance consistently across both image-level and pixel-level metrics (image AUROC/AP $91.4/95.7$, pixel AUROC/AUPRO $92.3/87.7$), followed by DINOv2.
For MVTec$\rightarrow$VisA, image-level detection is relatively close across backbones (AUROC $\approx 80$–$83$; AP $\approx 84$–$87$), while pixel-level segmentation shows clearer differences, with DINOv3 achieving the highest pixel AUROC/AUPRO ($96.0/89.1$) and SigLIP-L being notably lower in AUPRO ($81.8$).
Overall, SPG is compatible with both self-supervised ViTs and VLM-based visual encoders, while the choice of backbone can materially influence segmentation quality.

\begin{figure*}
    \centering
    \includegraphics[width=1.0\linewidth, trim=0 100 0 0 ]{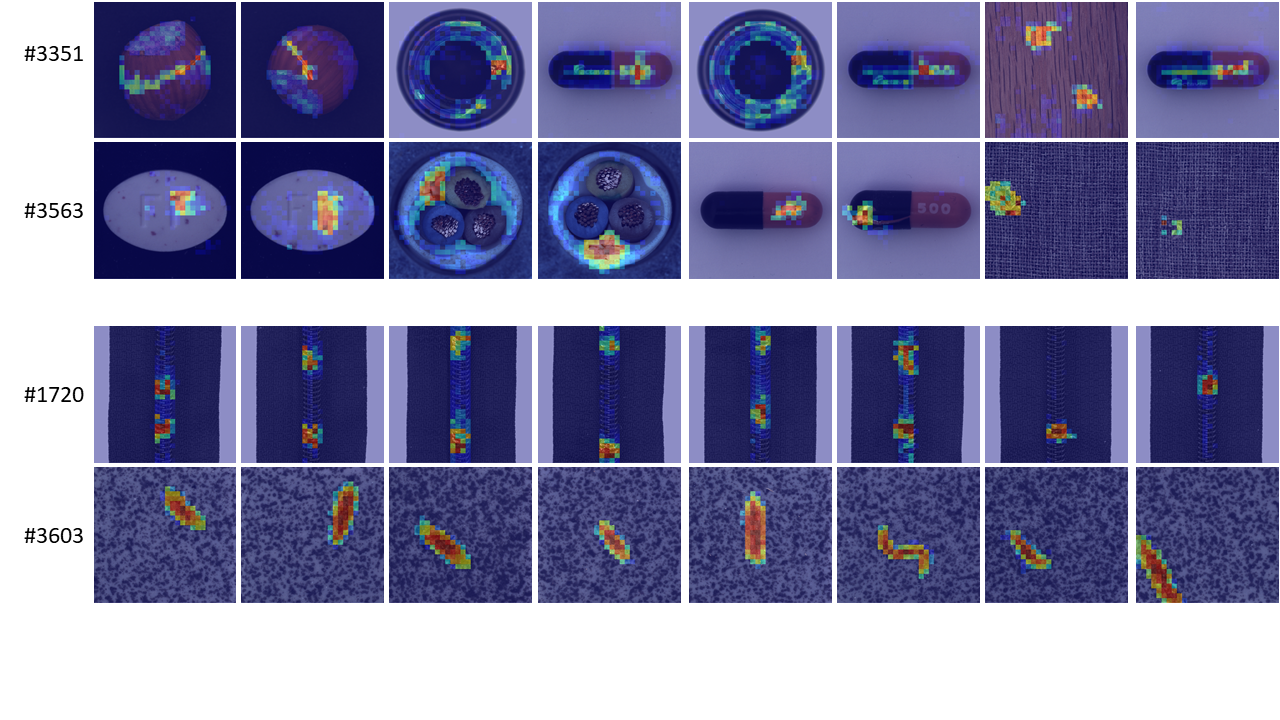}
    \caption{Qualitative interpretation of SAE dictionary atoms emphasized by the learned anomaly guide. We select representative atoms from the anomaly guide’s active set by choosing those with the largest learned anomaly coefficients. Each atom is visualized by retrieving top-activating patches from the auxiliary dataset and by overlaying an upsampled activation heatmap on the corresponding images for spatial context. This qualitative analysis illustrates that SPG’s anomaly criterion can be traced to a sparse subset of SAE atoms, some of which appear broadly activated across categories while others are more category-biased. }
    \label{fig:vis_atoms}
\end{figure*}

\subsection{Qualitative analysis of SAE atoms emphasized by the anomaly guide}
\label{subsec:vis_atoms}

In addition to quantitative evaluation, we provide a qualitative visualization of the SAE atoms emphasized by the learned anomaly criterion.

\noindent\textbf{Selecting atoms emphasized by the anomaly guide.}
Recall that the anomaly guide is defined by the SAE dictionary $D$ and the non-negative guide coefficients $\tilde{\bm{z}}_A$ (\cref{equ:guide}), with the active atom set $\mathcal{I}_A=\mathrm{supp}(\tilde{\bm{z}}_A)$ (\cref{equ:active_set}). We select a small number of atoms from $\mathcal{I}_A$ with the largest coefficients $\tilde{\bm{z}}_A[a]$ as representative atoms emphasized by the learned anomaly guide.

\noindent\textbf{Patch retrieval on the auxiliary dataset.}
To interpret what each selected atom represents in the frozen feature space, we visualize an atom by retrieving patches that strongly activate it on the auxiliary dataset used to train the SAE. For each auxiliary image $X_i$, we extract patch-token features $\bm{f}_{i,j}$ with the frozen encoder and compute SAE sparse codes $\tilde{\bm{z}}_{i,j}$ (\cref{equ:feature_extraction,equ:sae,equ:sae_encode}). We define the activation of the atom $a$ at patch $(i,j)$ as $\tilde{\bm{z}}_{i,j}[a]$ and retrieve the top-$K$ patches with the largest activation across the auxiliary dataset ($K=8$ in our visualizations). For context, we also visualize an atom-activation heatmap by placing $\tilde{\bm{z}}_{i,j}[a]$ on the patch grid and bilinearly upsampling it to the input resolution.

\noindent\textbf{Observation.}
\Cref{fig:vis_atoms} visualizes representative SAE atoms selected from the anomaly guide’s active set. 
Some atoms (top rows; e.g., \#3351 and \#3563) are activated by anomalous regions across multiple auxiliary categories, including cracks, missing parts, and other localized abnormal structures, suggesting that the learned anomaly guide exploits class-general anomaly cues in the shared SAE latent space.
In contrast, other atoms (bottom rows; e.g., \#1720 and \#3603) exhibit concentrated activation patterns that repeatedly co-occur with category-specific local structures, such as object parts or characteristic surface patterns, indicating more class-specific (category-biased) factors.
This qualitative dichotomy supports our motivation that sparse guide coefficients can be inspected to reveal what components the anomaly criterion relies on, rather than treating the decision anchor as a single opaque prompt embedding. 
Importantly, we treat these visualizations as evidence of interpretability at the level of individual dictionary atoms, without claiming semantic disentanglement of the entire SAE dictionary.

\section{Conclusion}

We proposed SPG for zero-shot anomaly detection and segmentation, which constructs normal/anomaly guide vectors by projecting sparse coefficients in an SAE latent space back to the frozen token feature space. SPG follows a two-stage procedure on an auxiliary dataset: training an SAE on patch tokens, then freezing it and optimizing only sparse guide coefficients to define the guide vectors as a sparse combination of dictionary atoms.

Across cross-dataset evaluations on MVTec AD and VisA, SPG achieves competitive image-level detection and strong pixel-level segmentation; with a DINOv3 backbone, it attains the best pixel-level AUROC among the compared methods in our study. 
Moreover, by inspecting the dictionary atoms emphasized by the learned guide coefficients, we observe that a sparse subset of SAE atoms appears to capture both category-general and category-specific factors within a shared SAE latent space.

{
    \small
    \bibliographystyle{ieeenat_fullname}
    \bibliography{main}
}

\clearpage
\appendix

\definecolor{cvprblue}{rgb}{0.21,0.49,0.74}

\def\confName{CVPR}
\def\confYear{2026}

\twocolumn[
\begin{center}
    {\Large\bfseries SPG: Sparse-Projected Guides with Sparse Autoencoders for Zero-Shot Anomaly Detection\par}
    \vspace{1em}
    {\large Supplementary Material\par}
    \vspace{2em}
\end{center}
]

\renewcommand{\thefigure}{S\arabic{figure}}
\renewcommand{\thetable}{S\arabic{table}}
\setcounter{figure}{0}
\setcounter{table}{0}

\section{Stage-2 Guide Learning with TopK Projection}\label{sec:supp_stage2_topk}
The main paper parameterizes Stage-2 guide coefficients using ReLU together with
an $\ell_1$ sparsity regularizer.
Here, we study an alternative based on explicit TopK sparsification to assess
how the choice of sparsity mechanism affects guide learning.

Let $\bar{\bm z}_N, \bar{\bm z}_A \in \mathbb{R}^C$ denote learnable coefficient parameters for the normal and anomalous guides, respectively.
We obtain $k_g$-sparse coefficients by
\begin{align}
    \tilde{\bm z}_N &= \mathrm{TopK}(\bar{\bm z}_N, k_g), \\
    \tilde{\bm z}_A &= \mathrm{TopK}(\bar{\bm z}_A, k_g),
\end{align}
where $\mathrm{TopK}(\cdot; k_g)$ keeps the $k_g$ largest entries and sets the remaining entries to zero.
Using the frozen SAE dictionary $\bm D$, the resulting guides are computed as $\bm g_N = \bm D \tilde{\bm z}_N$ and $\bm g_A = \bm D \tilde{\bm z}_A$.

\begin{figure}[t]
    \centering
    \includegraphics[width=\linewidth, trim=0 0 600 0, clip]{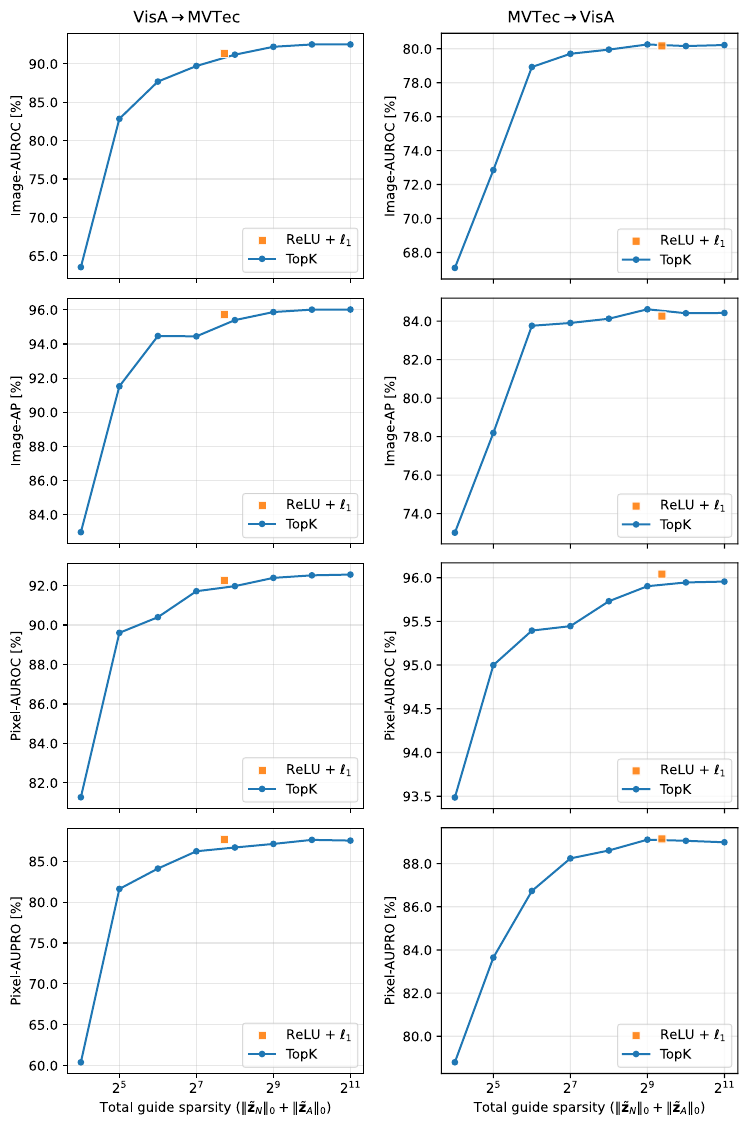}
    \caption{Stage-2 guide learning with explicit TopK sparsification. We replace the ReLU+$\ell_1$ parameterization in the main paper with $k_g$-sparse guide coefficients obtained by TopK projection, and plot performance against the total guide sparsity $\|\tilde{\bm z}_N\|_0+\|\tilde{\bm z}_A\|_0$. Orange markers denote the effective sparsity obtained under the ReLU+$\ell_1$ formulation in the main paper. In both transfer directions, performance improves as the sparsity budget increases, indicating that overly small TopK budgets over-constrain the guides.}
    \label{fig:sparsity}
\end{figure}

We sweep $k_g \in \{2^3,2^4,\ldots,2^{10}\}$ under the same cross-dataset protocol as in the main paper, namely VisA$\rightarrow$MVTec and MVTec$\rightarrow$VisA. 
For fair comparison with the ReLU+$\ell_1$ formulation, we plot performance against the total guide sparsity $\|\tilde{\bm z}_N\|_0+\|\tilde{\bm z}_A\|_0$.
For the TopK formulation, this corresponds to $2k_g$. 
We evaluate performance using Image AUROC, Image AP, Pixel AUROC, and Pixel AUPRO in~\cref{fig:sparsity}.

Performance improves consistently as the total TopK sparsity increases, and sufficiently large sparsity can match or slightly surpass the ReLU+$\ell_1$ formulation used in the main paper.
At the same time, the effective sparsity obtained under the ReLU+$\ell_1$ formulation differs substantially across transfer directions: the total number of non-zero normal/anomalous guide coefficients is 211 for VisA$\rightarrow$MVTec and 644 for MVTec$\rightarrow$VisA, respectively, for a dictionary width of 4096.
This observation suggests that different transfer directions may favor different effective sparsity levels.
In this sense, a fixed TopK constraint may be less flexible than the ReLU+$\ell_1$ formulation, which provides a softer and more continuous sparsity control.

\section{Per-Category Performance}
\label{sec:supp_per_category}

This section reports category-level results under cross-dataset transfer.
\Cref{tab:appendix_mvtecad_image_auroc,tab:appendix_mvtecad_image_ap,tab:appendix_mvtecad_pixel_auroc,tab:appendix_mvtecad_pixel_aupro}
summarize VisA$\rightarrow$MVTec AD results, including Image AUROC, Image AP,
Pixel AUROC, and Pixel AUPRO, respectively.
\Cref{tab:appendix_visa_image_auroc,tab:appendix_visa_image_ap,tab:appendix_visa_pixel_auroc,tab:appendix_visa_pixel_aupro}
report the corresponding results for MVTec AD$\rightarrow$VisA.
Entries marked with "--" indicate that category-wise values were not reported in the corresponding original paper.
When only dataset-level means were reported, we retain those mean values in the last row for reference.

\begin{table*}[t]
\centering
\caption{Category-wise image-level AUROC (\%) under the cross-dataset zero-shot setting (VisA $\rightarrow$ MVTec AD). Baseline results are taken from the original papers, and "--" denotes metrics not reported.}
\label{tab:appendix_mvtecad_image_auroc}
\scriptsize
\setlength{\tabcolsep}{3pt}
\resizebox{\textwidth}{!}{%
\begin{tabular}{lrrrrrrrr}
\toprule
Category & WinCLIP & AnomalyCLIP & AdaCLIP & VCP-CLIP & AA-CLIP & AdaptCLIP & SPG (OpenCLIP) & SPG (DINOv3) \\
\midrule
Bottle     & 99.2 & 89.3 & 94.4 & -- & -- & 90.7 & 78.7 & 99.3 \\
Cable      & 86.5 & 69.8 & 90.6 & -- & -- & 83.6 & 57.3 & 86.4 \\
Capsule    & 72.9 & 89.9 & 91.5 & -- & -- & 95.2 & 90.0 & 89.9 \\
Carpet     &100.0 &100.0 & 82.1 & -- & -- &100.0 & 92.3 & 97.8 \\
Grid       & 98.8 & 97.0 & 90.0 & -- & -- & 98.9 & 76.2 & 99.4 \\
Hazelnut   & 93.9 & 97.2 & 80.2 & -- & -- & 94.8 & 84.7 & 93.5 \\
Leather    &100.0 & 99.8 & 99.8 & -- & -- &100.0 & 97.2 &100.0 \\
Metal nut  & 97.1 & 93.6 & 83.5 & -- & -- & 93.7 & 69.8 & 97.5 \\
Pill       & 79.1 & 81.8 & 82.9 & -- & -- & 86.8 & 87.9 & 74.5 \\
Screw      & 83.3 & 81.1 & 87.0 & -- & -- & 85.9 & 69.4 & 86.9 \\
Tile       &100.0 &100.0 & 90.5 & -- & -- & 99.6 & 87.6 & 95.1 \\
Toothbrush & 87.5 & 84.7 & 93.6 & -- & -- & 86.4 & 62.2 & 84.7 \\
Transistor & 88.0 & 92.8 & 82.1 & -- & -- & 88.7 & 72.6 & 66.5 \\
Wood       & 99.4 & 96.8 & 98.3 & -- & -- & 98.7 & 73.5 & 99.1 \\
Zipper     & 91.5 & 98.5 & 91.5 & -- & -- & 99.3 & 90.8 & 99.7 \\
\midrule
\textbf{Mean} & 91.8 & 91.5 & 89.2 & 92.1 & 90.5 & 93.5 & 79.3 & 91.4 \\
\bottomrule
\end{tabular}}
\end{table*}

\begin{table*}[t]
\centering
\caption{Category-wise image-level AP (\%) under the cross-dataset zero-shot setting (VisA $\rightarrow$ MVTec AD). Baseline results are taken from the original papers, and "--" denotes metrics not reported.}
\label{tab:appendix_mvtecad_image_ap}
\scriptsize
\setlength{\tabcolsep}{3pt}
\resizebox{\textwidth}{!}{%
\begin{tabular}{lrrrrrrrr}
\toprule
Category & WinCLIP & AnomalyCLIP & AdaCLIP & VCP-CLIP & AA-CLIP & AdaptCLIP & SPG (OpenCLIP) & SPG (DINOv3) \\
\midrule
Bottle     & 99.8 & 97.0 & -- & -- & -- & 97.1 & 92.6 & 99.8 \\
Cable      & 91.2 & 81.4 & -- & -- & -- & 89.9 & 67.8 & 92.4 \\
Capsule    & 91.5 & 97.9 & -- & -- & -- & 99.0 & 97.8 & 98.0 \\
Carpet     &100.0 &100.0 & -- & -- & -- & 99.9 & 97.2 & 99.2 \\
Grid       & 99.6 & 99.1 & -- & -- & -- & 99.5 & 88.9 & 99.8 \\
Hazelnut   & 96.9 & 98.6 & -- & -- & -- & 97.0 & 90.5 & 97.1 \\
Leather    &100.0 & 99.9 & -- & -- & -- & 99.9 & 99.0 &100.0 \\
Metal nut  & 99.3 & 98.5 & -- & -- & -- & 98.3 & 91.9 & 99.4 \\
Pill       & 95.7 & 95.4 & -- & -- & -- & 97.0 & 97.6 & 94.8 \\
Screw      & 93.1 & 92.5 & -- & -- & -- & 93.8 & 88.0 & 95.7 \\
Tile       &100.0 &100.0 & -- & -- & -- & 99.7 & 90.7 & 98.4 \\
Toothbrush & 95.6 & 93.7 & -- & -- & -- & 92.5 & 84.7 & 94.7 \\
Transistor & 87.1 & 90.6 & -- & -- & -- & 87.1 & 69.2 & 67.0 \\
Wood       & 99.8 & 99.2 & -- & -- & -- & 99.5 & 91.0 & 99.7 \\
Zipper     & 97.5 & 99.6 & -- & -- & -- & 99.8 & 97.7 & 99.9 \\
\midrule
\textbf{Mean} & 96.5 & 96.2 & -- & 96.9 & -- & 96.7 & 89.6 & 95.7 \\
\bottomrule
\end{tabular}}
\end{table*}

\begin{table*}[t]
\centering
\caption{Category-wise pixel-level AUROC (\%) under the cross-dataset zero-shot setting (VisA $\rightarrow$ MVTec AD). Baseline results are taken from the original papers, and "--" denotes metrics not reported.}
\label{tab:appendix_mvtecad_pixel_auroc}
\scriptsize
\setlength{\tabcolsep}{3pt}
\resizebox{\textwidth}{!}{%
\begin{tabular}{lrrrrrrrr}
\toprule
Category & WinCLIP & AnomalyCLIP & AdaCLIP & VCP-CLIP & AA-CLIP & AdaptCLIP & SPG (OpenCLIP) & SPG (DINOv3) \\
\midrule
Bottle     & 89.5 & 90.4 & 90.4 & 94.1 & -- & 92.4 & 89.5 & 96.6 \\
Cable      & 77.0 & 78.9 & 79.8 & 78.6 & -- & 76.6 & 74.2 & 80.4 \\
Capsule    & 86.9 & 95.8 & 82.3 & 96.6 & -- & 95.8 & 93.7 & 95.6 \\
Carpet     & 95.4 & 98.8 & 97.3 & 99.6 & -- & 99.2 & 98.6 & 99.5 \\
Grid       & 82.2 & 97.3 & 96.9 & 97.9 & -- & 97.4 & 96.9 & 99.3 \\
Hazelnut   & 94.3 & 97.1 & 97.8 & 97.7 & -- & 97.6 & 94.6 & 97.7 \\
Leather    & 96.7 & 98.6 & 99.2 & 99.6 & -- & 99.1 & 99.1 & 99.7 \\
Metal nut  & 61.0 & 74.4 & 74.3 & 74.0 & -- & 76.5 & 77.0 & 87.0 \\
Pill       & 80.0 & 92.0 & 86.4 & 90.1 & -- & 89.9 & 87.0 & 85.9 \\
Screw      & 89.6 & 97.5 & 98.4 & 98.5 & -- & 97.9 & 98.2 & 98.0 \\
Tile       & 77.6 & 94.6 & 88.5 & 93.5 & -- & 95.8 & 91.8 & 89.2 \\
Toothbrush & 86.9 & 91.9 & 94.9 & 94.7 & -- & 87.7 & 93.2 & 94.2 \\
Transistor & 74.7 & 71.0 & 63.2 & 69.8 & -- & 69.3 & 66.1 & 65.9 \\
Wood       & 93.4 & 96.5 & 87.9 & 97.6 & -- & 96.9 & 93.4 & 95.9 \\
Zipper     & 91.6 & 91.4 & 93.8 & 98.1 & -- & 91.8 & 95.2 & 99.1 \\
\midrule
\textbf{Mean} & 85.1 & 91.1 & 88.7 & 92.0 & 91.9 & 90.9 & 89.9 & 92.3 \\
\bottomrule
\end{tabular}}
\end{table*}

\begin{table*}[t]
\centering
\caption{Category-wise pixel-level AUPRO (\%) under the cross-dataset zero-shot setting (VisA $\rightarrow$ MVTec AD). Baseline results are taken from the original papers, and "--" denotes metrics not reported.}
\label{tab:appendix_mvtecad_pixel_aupro}
\scriptsize
\setlength{\tabcolsep}{3pt}
\resizebox{\textwidth}{!}{%
\begin{tabular}{lrrrrrrrr}
\toprule
Category & WinCLIP & AnomalyCLIP & AdaCLIP & VCP-CLIP & AA-CLIP & AdaptCLIP & SPG (OpenCLIP) & SPG (DINOv3) \\
\midrule
Bottle     & 76.4 & 80.9 & -- & 88.5 & -- & -- & 72.4 & 91.4 \\
Cable      & 42.9 & 64.4 & -- & 67.7 & -- & -- & 39.4 & 73.7 \\
Capsule    & 62.1 & 87.2 & -- & 94.0 & -- & -- & 70.5 & 86.0 \\
Carpet     & 84.1 & 90.1 & -- & 98.4 & -- & -- & 85.7 & 98.2 \\
Grid       & 57.0 & 75.6 & -- & 92.4 & -- & -- & 75.6 & 94.7 \\
Hazelnut   & 81.6 & 92.4 & -- & 84.9 & -- & -- & 78.1 & 86.4 \\
Leather    & 91.1 & 92.2 & -- & 99.2 & -- & -- & 91.2 & 98.5 \\
Metal nut  & 31.8 & 71.0 & -- & 73.4 & -- & -- & 67.4 & 84.1 \\
Pill       & 65.0 & 88.2 & -- & 94.9 & -- & -- & 80.3 & 89.6 \\
Screw      & 68.5 & 88.0 & -- & 93.2 & -- & -- & 81.6 & 86.3 \\
Tile       & 51.2 & 87.6 & -- & 90.6 & -- & -- & 69.3 & 85.2 \\
Toothbrush & 67.7 & 88.5 & -- & 87.8 & -- & -- & 83.6 & 88.6 \\
Transistor & 43.4 & 58.1 & -- & 56.1 & -- & -- & 45.0 & 60.8 \\
Wood       & 74.1 & 91.2 & -- & 95.7 & -- & -- & 82.5 & 95.4 \\
Zipper     & 71.7 & 65.3 & -- & 92.2 & -- & -- & 55.6 & 96.0 \\
\midrule
\textbf{Mean} & 64.6 & 81.4 & -- & 87.3 & -- & -- & 71.9 & 87.7 \\
\bottomrule
\end{tabular}}
\end{table*}

\begin{table*}[t]
\centering
\caption{Category-wise image-level AUROC (\%) under the cross-dataset zero-shot setting (MVTec AD $\rightarrow$ VisA). Baseline results are taken from the original papers, and "--" denotes metrics not reported.}
\label{tab:appendix_visa_image_auroc}
\scriptsize
\setlength{\tabcolsep}{3pt}
\resizebox{\textwidth}{!}{%
\begin{tabular}{lrrrrrrrr}
\toprule
Category & WinCLIP & AnomalyCLIP & AdaCLIP & VCP-CLIP & AA-CLIP & AdaptCLIP & SPG (OpenCLIP) & SPG (DINOv3) \\
\midrule
Candle     & 95.4 & 79.3 & 96.0 & -- & -- & 87.4 & 84.5 & 74.7 \\
Capsules   & 85.0 & 81.5 & 85.1 & -- & -- & 93.9 & 92.4 & 95.8 \\
Cashew     & 92.1 & 76.3 & 91.8 & -- & -- & 86.1 & 84.3 & 70.2 \\
Chewinggum & 96.5 & 97.4 & 96.4 & -- & -- & 96.7 & 96.2 & 97.7 \\
Fryum      & 80.3 & 93.0 & 93.0 & -- & -- & 91.5 & 91.9 & 94.9 \\
Macaroni1  & 76.2 & 87.2 & 91.6 & -- & -- & 83.3 & 77.3 & 72.7 \\
Macaroni2  & 63.7 & 73.4 & 64.1 & -- & -- & 69.9 & 61.2 & 64.2 \\
PCB1       & 73.6 & 85.4 & 81.1 & -- & -- & 84.1 & 75.6 & 65.9 \\
PCB2       & 51.2 & 62.2 & 75.3 & -- & -- & 64.6 & 69.9 & 81.8 \\
PCB3       & 73.4 & 62.7 & 64.7 & -- & -- & 65.3 & 66.4 & 53.3 \\
PCB4       & 79.6 & 93.9 & 93.4 & -- & -- & 97.7 & 93.1 & 91.1 \\
Pipe fryum & 69.7 & 92.4 & 96.6 & -- & -- & 96.7 & 98.0 & 99.6 \\
\midrule
\textbf{Mean} & 78.1 & 82.1 & 85.8 & 84.6 & 82.6 & 84.8 & 82.6 & 80.2\\
\bottomrule
\end{tabular}}
\end{table*}

\begin{table*}[t]
\centering
\caption{Category-wise image-level AP (\%) under the cross-dataset zero-shot setting (MVTec AD $\rightarrow$ VisA). Baseline results are taken from the original papers, and "--" denotes metrics not reported.}
\label{tab:appendix_visa_image_ap}
\scriptsize
\setlength{\tabcolsep}{3pt}
\resizebox{\textwidth}{!}{%
\begin{tabular}{lrrrrrrrr}
\toprule
Category & WinCLIP & AnomalyCLIP & AdaCLIP & VCP-CLIP & AA-CLIP & AdaptCLIP & SPG (OpenCLIP) & SPG (DINOv3) \\
\midrule
Candle     & 95.8 & 81.1 & -- & -- & -- & 90.5 & 88.0 & 81.2 \\
Capsules   & 90.9 & 88.7 & -- & -- & -- & 96.8 & 96.3 & 98.1 \\
Cashew     & 96.4 & 89.4 & -- & -- & -- & 94.0 & 93.0 & 84.0 \\
Chewinggum & 98.6 & 98.9 & -- & -- & -- & 98.7 & 98.1 & 99.1 \\
Fryum      & 90.1 & 96.8 & -- & -- & -- & 95.9 & 96.4 & 97.7 \\
Macaroni1  & 75.8 & 86.0 & -- & -- & -- & 85.1 & 82.5 & 77.8 \\
Macaroni2  & 60.3 & 72.1 & -- & -- & -- & 69.2 & 61.2 & 63.8 \\
PCB1       & 78.4 & 87.0 & -- & -- & -- & 86.1 & 80.7 & 73.8 \\
PCB2       & 49.2 & 64.3 & -- & -- & -- & 67.8 & 70.7 & 85.6 \\
PCB3       & 76.5 & 70.0 & -- & -- & -- & 72.0 & 69.7 & 58.3 \\
PCB4       & 77.7 & 94.4 & -- & -- & -- & 97.5 & 93.0 & 92.0 \\
Pipe fryum & 82.3 & 96.3 & -- & -- & -- & 98.0 & 99.0 & 99.8 \\
\midrule
\textbf{Mean} & 81.2 & 85.4 & -- & 87.6 & -- & 87.6 & 85.7 & 84.3 \\
\bottomrule
\end{tabular}}
\end{table*}

\clearpage

\begin{table*}[t]
\centering
\caption{Category-wise pixel-level AUROC (\%) under the cross-dataset zero-shot setting (MVTec AD $\rightarrow$ VisA). Baseline results are taken from the original papers, and "--" denotes metrics not reported.}
\label{tab:appendix_visa_pixel_auroc}
\scriptsize
\setlength{\tabcolsep}{3pt}
\resizebox{\textwidth}{!}{%
\begin{tabular}{lrrrrrrrr}
\toprule
Category & WinCLIP & AnomalyCLIP & AdaCLIP & VCP-CLIP & AA-CLIP & AdaptCLIP & SPG (OpenCLIP) & SPG (DINOv3) \\
\midrule
Candle     & 88.9 & 98.8 & 98.9 & 99.2 & -- & 98.7 & 98.8 & 98.9 \\
Capsules   & 81.6 & 95.0 & 98.6 & 98.7 & -- & 94.2 & 95.6 & 99.4 \\
Cashew     & 84.7 & 93.8 & 95.9 & 93.1 & -- & 93.4 & 91.7 & 96.2 \\
Chewinggum & 93.3 & 99.3 & 99.6 & 99.5 & -- & 99.5 & 99.5 & 99.5 \\
Fryum      & 88.5 & 94.6 & 94.4 & 94.6 & -- & 95.3 & 92.0 & 94.7 \\
Macaroni1  & 70.9 & 98.3 & 99.5 & 99.6 & -- & 98.1 & 98.4 & 98.4 \\
Macaroni2  & 59.3 & 97.6 & 98.8 & 98.7 & -- & 98.2 & 98.1 & 98.1 \\
PCB1       & 61.2 & 94.1 & 93.7 & 92.1 & -- & 96.2 & 92.4 & 91.3 \\
PCB2       & 71.6 & 92.4 & 84.3 & 92.1 & -- & 93.0 & 89.5 & 94.6 \\
PCB3       & 85.3 & 88.4 & 91.8 & 89.2 & -- & 87.9 & 90.8 & 89.1 \\
PCB4       & 94.4 & 95.7 & 96.1 & 95.6 & -- & 95.3 & 95.3 & 95.1 \\
Pipe fryum & 75.4 & 98.2 & 94.6 & 96.4 & -- & 98.4 & 94.0 & 97.2 \\
\midrule
\textbf{Mean} & 79.6 & 95.5 & 95.5 & 95.7 & 95.5 & 95.7 & 94.7 & 96.0 \\
\bottomrule
\end{tabular}}
\end{table*}

\begin{table*}[t]
\centering
\caption{Category-wise pixel-level AUPRO (\%) under the cross-dataset zero-shot setting (MVTec AD $\rightarrow$ VisA). Baseline results are taken from the original papers, and "--" denotes metrics not reported.}
\label{tab:appendix_visa_pixel_aupro}
\scriptsize
\setlength{\tabcolsep}{3pt}
\resizebox{\textwidth}{!}{%
\begin{tabular}{lrrrrrrrr}
\toprule
Category & WinCLIP & AnomalyCLIP & AdaCLIP & VCP-CLIP & AA-CLIP & AdaptCLIP & SPG (OpenCLIP) & SPG (DINOv3) \\
\midrule
Candle     & 83.5 & 96.2 & -- & 96.2 & -- & -- & 96.5 & 89.9 \\
Capsules   & 35.3 & 78.5 & -- & 91.2 & -- & -- & 81.8 & 98.6 \\
Cashew     & 76.4 & 91.6 & -- & 95.6 & -- & -- & 93.8 & 88.7 \\
Chewinggum & 70.4 & 91.2 & -- & 92.2 & -- & -- & 94.8 & 92.8 \\
Fryum      & 77.4 & 86.8 & -- & 92.2 & -- & -- & 90.6 & 92.9 \\
Macaroni1  & 34.3 & 89.8 & -- & 97.5 & -- & -- & 90.2 & 91.7 \\
Macaroni2  & 21.4 & 84.2 & -- & 90.5 & -- & -- & 83.5 & 91.1 \\
PCB1       & 26.3 & 81.7 & -- & 88.1 & -- & -- & 83.7 & 86.4 \\
PCB2       & 37.2 & 78.9 & -- & 79.1 & -- & -- & 72.3 & 80.2 \\
PCB3       & 56.1 & 77.1 & -- & 78.9 & -- & -- & 78.2 & 73.1 \\
PCB4       & 80.4 & 91.3 & -- & 89.7 & -- & -- & 84.7 & 86.6 \\
Pipe fryum & 82.3 & 96.8 & -- & 96.5 & -- & -- & 95.2 & 97.7 \\
\midrule
\textbf{Mean} & 56.8 & 87.0 & -- & 90.7 & -- & -- & 87.1 & 89.1 \\
\bottomrule
\end{tabular}}
\end{table*}

\end{document}